%% file: main.tex
\documentclass[11pt]{article}

\usepackage[preprint]{acl}

\usepackage{times}
\usepackage{latexsym}
\usepackage{microtype}
\usepackage{graphicx}
\usepackage{subcaption}
\usepackage{booktabs} % for professional tables
\usepackage{amsmath}
\usepackage{amssymb}
\usepackage{mathtools}
\usepackage{amsthm}
\usepackage{hyperref}
\usepackage{algorithm}
\usepackage{algorithmic}

\usepackage{hyperref}
\usepackage[capitalize,noabbrev]{cleveref}

\usepackage[textsize=tiny]{todonotes}

\usepackage[T1]{fontenc}
\usepackage[utf8]{inputenc}

\usepackage{microtype}

\usepackage{inconsolata}

\usepackage{graphicx}

\title{POTracker: Optimizing Large Language Models for Standard-Compliant Power Outage Report Generation}

\author{
Hung Phan \\
Iowa State University \\
\texttt{hungphd@iastate.edu}
\And
Aniroop Naladala \\
Iowa State University \\
\texttt{aniroopn@iastate.edu}
\And
Avanindra Dubey \\
Iowa State University \\
\texttt{avid421@iastate.edu}
\AND
Supriya Chinthavali \\
Oak Ridge Nat. Laboratory \\
\texttt{chinthavalis@ornl.gov}
\And
Dalton Lunga \\
Oak Ridge Nat. Laboratory \\
\texttt{lungadd@ornl.gov}
\And
Ali Jannesari \\
Iowa State University \\
\texttt{jannesar@iastate.edu}
}

\begin{document}
\maketitle
\input{Abstract}
\input{Introduction}
\input{MotivationExample}
\input{Background}

\input{ProposedPipelines}

\input{Experiment}
\input{RelatedWork}
\input{Discussion}

\input{Conclusion}
\newpage
\input{Acknowledgement}

% \newpage
\bibliography{refs}
\onecolumn
\appendix

\input{Appendix}

\end{document}

%% file: Abstract.tex
\begin{abstract}

Recent large language models (LLMs) are good at general text generation, but it is still hard to use them for domain-specific data generation because the output must follow strict format and structure rules. Different from open-ended tasks like question answering or translation, domain-specific generation must be both correct in meaning and also follow existing guidelines and standards. In this work, we study the nationwide interoperability problem of utility power outage reports in the United States. In practice, outage reports need to be machine-readable (e.g., JSON or XML) and must strictly follow requirements from energy-sector regulatory bodies. To handle this, we propose POTracker, an optimized LLM for power outage report generation. We fine-tune Qwen2.5-7B-Instruct with our proposed objective. The key contribution is a new loss function, $POTracker_{Loss}$, that considers both textual similarity and structure (tag) similarity between the generated report and the ground-truth report. We evaluate POTracker on a dataset of 1,000 power outage reports and compare it with five well-known fine-tuning methods and one rule-based XML conversion method. Results show that POTracker performs better than other fine-tuning approaches, improving overall accuracy by up to 51\% and reaching 86.47\% structural accuracy for generated power outage reports. In addition, we conduct a human study to assess the quality of the ground-truth standard reports, where domain experts assign the generated labels an average score of \textbf{4.03} on a 0--5 scale.

\end{abstract} 

%% file: Introduction.tex
\section{Introduction}
Large language models (LLMs) have demonstrated remarkable capabilities across a variety of natural language processing (NLP) tasks, including summarization \cite{zhang2023benchmarkinglargelanguagemodels}, question answering \cite{kamalloo2023evaluatingopendomainquestionanswering}, and machine translation \cite{zhu2024multilingualmachinetranslationlarge}. These models are increasingly being adopted in real-world applications due to their generalization abilities and fluent text generation. However, applying LLMs to domain-specific generation tasks remains a significant challenge \cite{ling2024domainspecializationkeymake}. Unlike general-purpose applications, many domain-specific tasks require the output to be not only semantically accurate but also to strictly follow formal conventions defined by industry or institutional standards. 

We identify two major challenges in applying Large Language Models (LLMs) to solve domain-specific problems. First, as the term suggests, these problems require the integration of knowledge from specialized resources at the industry or institutional level. This makes it difficult for general-purpose LLMs to provide high-quality answers, especially in terms of semantic accuracy. For instance, \cite{phan2024examininglongcontextlargelanguage} shows that in the domain of environmental assessment, state-of-the-art models such as GPT-4 \cite{openai2024gpt4technicalreport} and Claude \cite{anthropic2024claude3} achieved low accuracy when answering open-ended questions sourced from domain experts. Second, the outputs users expect from LLMs are not limited to textual responses but often extend to other data formats, depending on specific requirements. For example, \cite{hollmann2023large} optimized LLMs for feature generation in machine learning tasks. Other studies have focused on enhancing LLMs for machine learning interpretability \cite{bordt2024datasciencellmsinterpretable}, code generation \cite{nascimento2024llm4dsevaluatinglargelanguage}, and code performance optimization \cite{mahmud2025autoparllmgnnguidedcontextgeneration}. Compared to general text generation tasks, these applications demand not only semantic accuracy but also syntactic validity. Consequently, a second challenge arises: the need for more stringent evaluation policies to assess output quality under domain-specific and format-specific constraints.

In this project, our domain for conducting a power outage report generation is the Outage Data Initiative Nationwide (ODIN) benchmark \cite{ODIN2023}, provided by the Oak Ridge National Laboratory (ORNL).  Its primary goal is to establish a comprehensive digital reporting standard for power outage data, enabling utilities and stakeholders to exchange information seamlessly. This initiative aims to enhance restoration efforts, reliability, risk mitigation, and emergency response by providing standardized and real-time outage information. A central feature of ODIN is its delivery of real-time power outage data at the county level. This information is accessible via the Open Energy Data Portal, providing stakeholders with up-to-date insights into outages across various regions. By leveraging this data, emergency responders, utility providers, and other key entities can enhance their situational awareness and better coordinate their responses to power disruptions. Each data point provided in the ODIN dataset needs to follow standard of Common Information Model (CIM) IEC 61968-3 \cite{IEC61968-3:2021}. Various defined templates of ODIN data points can be found in the ODIN documentation \cite{ODIN_Developer_Guide}. Users face several challenges when providing input to the ODIN system. Often, the input is submitted in unstructured or free-form formats, including non-standard XML and JSON files, as well as free-text descriptions. This lack of standardization complicates data processing and integration. 
% Additionally, data scientists are typically required to write custom code to extract statistics and insights from datasets within the ODIN benchmark, adding further overhead to the data analysis process.

To overcome this challenge, we address the following important task in the context of power outage reporting: Power Outage Report Generation. This task involves transforming loosely structured or noisy XML documents, collected from users, into canonical representations as defined by ORNL documentation. 
% The generation task, on the other hand, requires producing valid and informative XML reports from either natural language descriptions or incomplete metadata, provided from all level of users from data collectors to data scientists. 
% The evaluation phases of these tasks will test the model's ability to balance semantic correctness with structural, an area where most current LLMs are not explicitly trained. 
In this work, we consider the well-known Close model, GPT-5.2 \cite{openai_gpt52_introducing_2025}, for generating labeled power outage reports. We propose $POTracker_{metric}$ score for meassuring the quality of the generated report given a synthetic report as label by GPT-5.2 with respoect to both context and tag similarities. By experiment, we show whether or not our proposed $POTracker_{Loss}$ can significantly improve Qwen-2.5-7b-Instruct in power outage reports' generation against multiple baselines, including a rule-based approach and other fine-tuning optimization approaches. We propose following contribution. Supplemental material is provided at here\footnote{https://tinyurl.com/3tat5sy5}.
% Specifically, we apply (1) test-time computation enhancements, (2) supervised fine-tuning (SFT), and (3) Direct Preference Optimization (DPO). Each of these techniques aims to align model outputs more closely with the strict structural and content requirements imposed by the target domain. We construct two evaluation datasets specifically tailored for Power Outage standardization and generation. These datasets were curated through a semi-automated pipeline: initial annotations were generated using GPT-4, followed by manual validation by human annotators. This process resulted in high-quality ground truth data for evaluating model performance in a controlled and repeatable way. Experimental results show that our optimization strategies lead to significant improvements over baseline prompt-only approaches. Specifically, we achieve up to 7\% relative improvement in XML standardization accuracy and 8\% improvement in Power Outage generation quality compared to unoptimized open LLMs. These gains demonstrate the potential of tailored optimization techniques for adapting general-purpose models to structure-sensitive, domain-specific tasks. Our contributions are as follows:

\begin{enumerate}
    \item \textbf{Metric.} We design $POTracker_{Metric}$, a formula for measuring the similarity between generated and predicted power outage reports with the consideration of both textual and structural correctness.
    \item \textbf{Custom-Loss Fine Tuning}. We propose $POTracker_{Loss}$, a specific loss function with the integration to the traditional cross entropy loss function to allow LLMs better generating power outage report
    % \item We introduce two high-quality datasets by a semi-automatic approach that takes advantage of both strong LLMs generation and human knowledge for data initialization and generation.
    \item \textbf{Evaluation}. We implemented multiple strategies for power outage report generation as baselines for comparison with POTracker in power outage report generation. We show that POTracker can outperform baselines in terms of accuracy and performance.
\end{enumerate}

Next sections are provided as following. The Motivation Example and Background sections provide examples and important concepts/ design selection of our work. The next section, Approach, introduces our proposed fine-tuning with our designed custom loss function, along with alternative fine-tuning strategies we use as the baselines. Experiment introduces our evaluation's configuration and results. Final sections are Related Work, Discussion and Conclusion.
% This work highlights the importance of domain-specific evaluation and adaptation for LLM deployment in real-world, safety-critical settings. We release our datasets and code to facilitate further research in structure-constrained text generation.
% \input{tables/tbl_configuration}
% \input{tables/tbl_accuracy_v2}

%% file: MotivationExample.tex
\section{Motivation Example}

\begin{figure}[htbp]
  \centering
  \includegraphics[width=\linewidth]{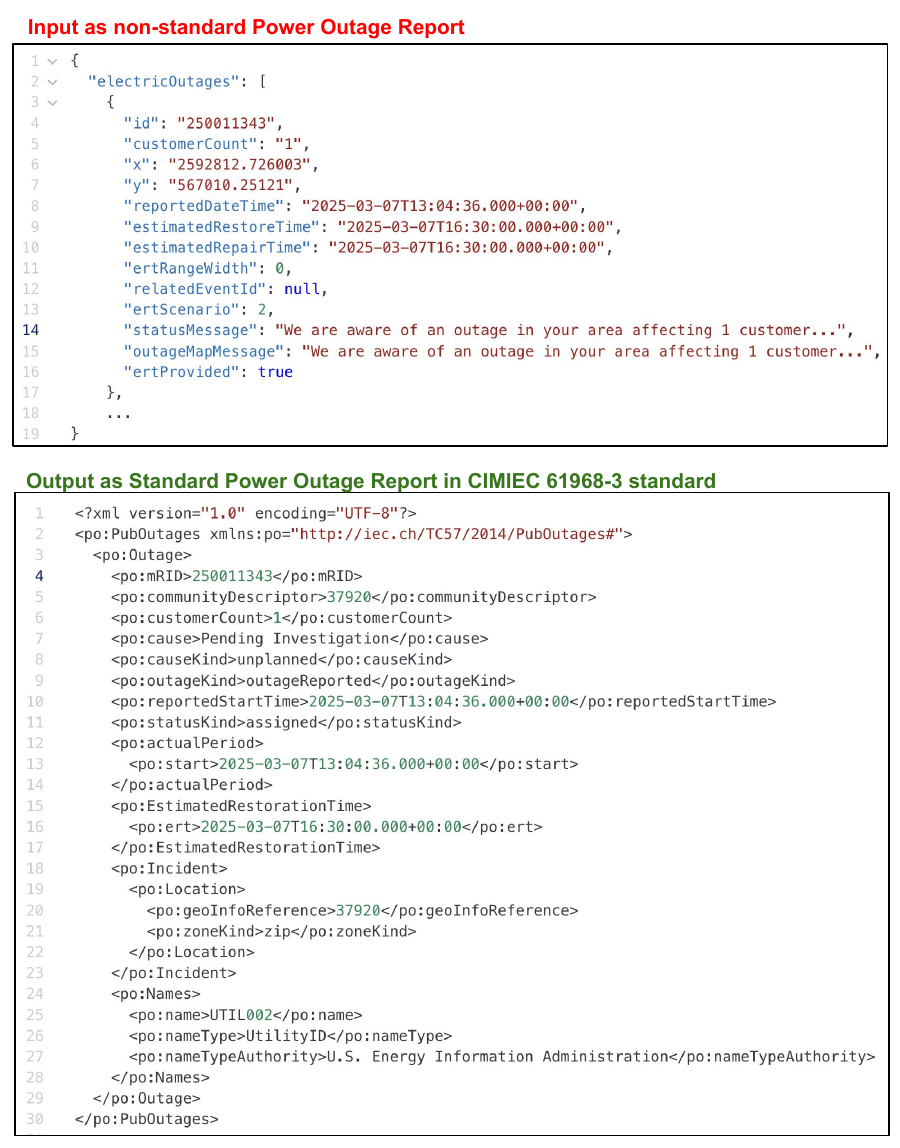}
  \caption{Examples of inputs and output of POTracker Standardization and Generation .}
  \label{fig:outage-zip-37920}
\end{figure}

% \textbf{Power Outage Report Standardization.} 
An example of the power outage generation process is shown in Figure~\ref{fig:outage-zip-37920}. In this scenario, the input—provided by data collectors—contains a list of electric outage reports by region. For each region, various types of information may be included, such as the number of affected customers, the outage timeline, and a status message. It is important to note that outage reports may differ in the properties they contain. For instance, some reports include a \texttt{relatedEventId} field that links to other related outage events. The output, formatted according to ODIN templates, adheres to a more restrictive and unified schema for all electric outage reports. To transform nonstandard inputs into this standardized format, domain experts are often required to manually map properties and values to the target template specified in the ODIN documentation~\cite{ODIN_Developer_Guide}.

% \textbf{Power Outage Report Generation.} While XML remains a popular input format, users interacting with data science tools are increasingly interested in using natural language queries\cite{mohammadjafari2025naturallanguagesqlreview}. An example of such input is demonstrated in the pair labeled Input 2 and Output in Figure~\ref{fig:outage-zip-37920}. In this case, the input query specifies parameters such as a ZIP code and a range for the number of affected customers. The expected output is a standardized XML file conforming to the ODIN specification, generated through two stages of filtering. Although this use case shares similarities with queries written in database-retrieval languages such as SQL~\cite{chamberlin1996history}, natural language queries cannot be manually processed due to the variability in how users express properties and values. Therefore, we aim to build a system capable of automatically interpreting natural language queries and generating corresponding power outage reports in the ODIN-compliant XML format.

%% file: Background.tex
\begin{figure*}[h]
\includegraphics[width=\linewidth]{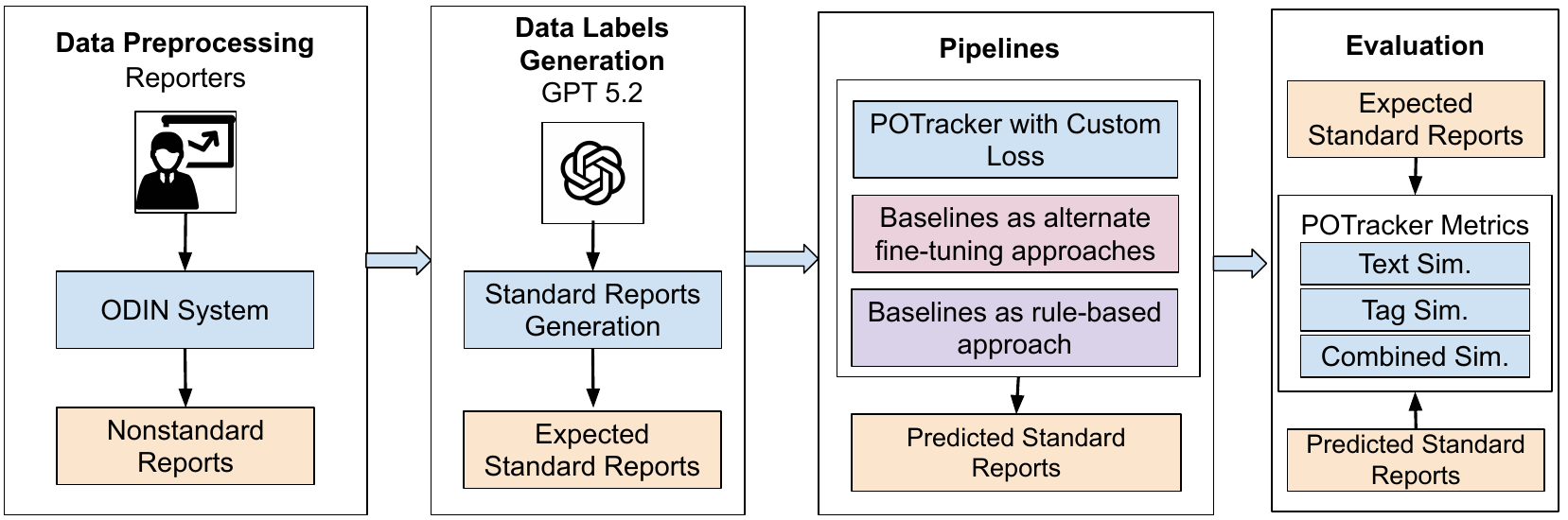}
\caption{Architecture Overview of POTracker.}
\label{fig:potracker_overview}

\end{figure*}

\section{Background}
In this section, we describe key concepts used throughout the paper.

\textbf{Power Outage Tracker.} In the context of this paper, the Power Outage Tracker refers to an application provided by ORNL through the ODIN system. Given inputs such as outage reports submitted by individuals across the United States, the application generates a standardized outage report following the CIM-IEC-61968-3 specification~\cite{IEC61968-3:2021}. This tool allows developers to submit various forms of input, including incomplete XML or JSON reports, as well as natural language descriptions. The application is integrated into the chatbot interface of the ODIN live map, accessible at this location\footnote{\url{https://odin.ornl.gov/outagemap/index.html}}.

\textbf{Selected Dataset.} We use the dataset of \textbf{1000} entities of power outage reports in USA collected from March, 2024. These entities are provided in a non-standard format. We split \textbf{800} entities for training and \textbf{100} for validation during fine-tuning. The other 100 entities was used for testing the performance of models.

\textbf{Selected LLMs.} For standard reports as labeled data, we use GPT 5.2 \cite{openai_gpt52_introducing_2025} for generating a standard power outage report following the CIM-IEC-61968-3 standard. We build our pipeline as a power outage report prediction application based on the fine-tuning process of the well-known and compact LLM Qwen-2.5-7b-Instruct\footnote{https://huggingface.co/Qwen/Qwen2.5-7B-Instruct}.

% \textbf{SambaNova System.} Selecting the appropriate open LLMs is critical, especially given that over 207,000 models were listed as "trending" on the Hugging Face platform as of April 2025~\cite{huggingface2024models}. For LLM fine-tuning, we use the SambaNova platform~\cite{prabhakar2024sambanova} to optimize open models for the Power Outage Tracker application. Specifically, we fine-tune the top five most frequently used open LLMs within ORNL’s SambaNova system for the task of generating standardized power outage reports.

%% file: ProposedPipelines.tex
\section{Approach}
In this section, we describe the modules provided for the POTracker application, shown in Figure \ref{fig:potracker_overview}. 

\subsection{Data Preprocessing}
In this module, the input data is preprocessed to extract meaningful information related to power outages. ODIN’s data collection process is designed to ingest outage reports from a wide range of sources, including utility companies, emergency response agencies, and even the general public. These inputs reflect real-time or near-real-time conditions across different regions in the United States. Because contributors operate at different technical levels, the incoming data varies widely in format—ranging from well-structured XML and JSON files to unstructured natural language descriptions submitted via forms or chat interfaces. Utility companies often send data directly from their Outage Management Systems (OMS) or Geographic Information Systems (GIS), while emergency responders may relay information manually or through partner platforms. Public contributions may include citizen reports submitted through local portals or integrated communication channels. ODIN accommodates this heterogeneity using a flexible interface that supports both automated API submissions and manual uploads. Each report, regardless of format, is parsed and preprocessed to extract key metadata such as location identifiers, outage start time, number of affected customers, and restoration estimates. The data is then queued for validation and transformed into a standardized schema. We refer to this level of raw input data as \textbf{nonstandard (power outage) reports}.

\subsection{Data Label Generation}
One significant challenge is that, while there are many datasets containing users' power outage reports, there is no readily available dataset of expected outputs in the CIM-IEC-61968-3 standard format as defined by the ODIN documentation~\cite{IEC61968-3:2021}. Although these standardized datasets—referred to as \textit{standard reports}—are critical for guiding LLM optimization, generating them manually is resource-intensive due to the dynamic nature of outage reports and the sheer volume of user-submitted data each year. To address this,  we employ closed LLMs guided by tailored prompts to generate large-scale standard reports from nonstandard inputs. Closed LLMs, which require access fees and API tokens, are known for their high-quality generative capabilities. In our project, We prompt GPT-5.2 to convert nonstandard reports into standard reports based on the requirements outlined in the ODIN documentation. The resulting outputs serve as candidate labels for further validation. We provide the prompt of data label generation in the replication package of this project.

\subsection{Evaluation.} 
The target and predicted objects for our problem don't only represent textual content but also structured content. We focus on the problem of comparing standard XMLs. To the best of our knowledge, there is no prior work provided a module for evaluating the quality of XML files generated by LLMs, since LLM-based XML generation systems are mainly at idealizing phase \cite{Zahid_Joudar_2024}. We design \textbf{$POTracker_{Metric}$}, a metric for calculating the similarities between XML files as predicted and expected results. Overall, the $POTracker_{Metric}$ function is designed as a combination formula of the following aspects.

\paragraph{Text Similarity.} 
The first key aspect of XML comparison is \textit{text similarity}, which evaluates how closely the textual values of corresponding XML nodes match between two documents. This is especially useful when validating whether the transformed or generated document preserves the intended semantics from the original. Text similarity is critical in high-stakes domains like outage reporting, where changes to specific fields—such as customer counts or timestamps—could lead to misinformation or operational errors. Accurately capturing these content differences ensures that outputs generated by models or pipelines maintain their factual integrity across revisions. We measure the Text Similarity score between two given reports using cosine similarity \cite{manning2008ir}.

\paragraph{Tag Structure Similarity.} 
Another fundamental dimension is \textit{tag structure similarity}, which measures how well the XML structure aligns between two documents. This includes the hierarchical arrangement of elements, the presence of required attributes, and the ordering of nodes. This is especially relevant for schema validation, where even small structural inconsistencies—such as missing wrapper elements or reordered tags—can cause downstream parsing or validation to fail. Tag structure comparison is therefore important for ensuring interoperability between systems that rely on standards such as CIM-IEC-61968-3. To quantify tag similarity between standard-format power outage reports, we use the sequence-matching approach implemented in Python's \texttt{difflib.SequenceMatcher} library, which identifies the longest contiguous matching blocks between two sequences and returns a normalized similarity score in the range $[0,1]$~\cite{python_difflib}.

The $POTracker_{Metic}$ function is calculated as the combination of Content Similarity and Tag Structure Similarity by following:

\begin{align}
\text{$POTracker\_{Metric}$} =\ & \alpha \cdot \text{Text\_Similarity} \nonumber \\
& + (1 - \alpha) \cdot \text{Tag\_Similarity}
\label{eq:overall_similarity}
\end{align}

In formula \ref{eq:overall_similarity}, $\alpha \in [0,1]$ is a weighting factor that reflects the importance of structural integrity relative to content accuracy. 
% We use xmldiff \cite{xmldiff} to calculate component scores. 
By default, we set the $\alpha = 0.5$.

\subsection{Pipelines}
In the era of large language models (LLMs), a key challenge is adapting a general-purpose model to new knowledge and domain-specific requirements. A common solution is fine-tuning, where the model is trained on paired examples—each sample includes a prompt and an expected response in the required format—so it learns to follow task instructions and generate compliant outputs \citep{wei2022finetuned}. In practice, many modern pipelines use parameter-efficient fine-tuning (PEFT), which keeps the base model fixed and learns lightweight task-specific modules (e.g., adapters or low-rank updates). This approach produces small adapter weights that can be attached to the original LLM to steer domain-specific generation \citep{pmlr-v97-houlsby19a,hu2021lora}. In our work, in addition to our proposed pipelines, we evaluate two baseline branches. First, we implement several established fine-tuning variants and use their learned adapters to assess the quality of power report generation. Second, we implement a simpler rule-based conversion method for power report generation and compare its performance against learning-based approaches. We show baselines' algorithms in the Appendix.

\subsubsection{POTracker with Custom Loss Function.}
\input{algms/alg_potracker_customloss}
We design our proposed custom loss function in Algorithm \ref{alg:potracker-customloss}. Given the fact that cross-entropy loss is frequently used in many fine-tuning pipelines,  we train the model with standard token-level cross-entropy so it can learns to predict the correct XML tokens. To enhance the ability of fine-tuning in generating good power outage reports in terms of XML format we also add a sequence-level quality signal that measures whether the entire generated XML looks correct (Line 5). For each mini-batch, we form a greedy predicted XML (from parameter logits $Z$, decode it and the gold XML into text, and compute a single similarity score that reflects both surface text match and XML tag/structure consistency. This score is then converted into a penalty and added to the cross-entropy objective with a weighting factor $\lambda$. In our setting, we set this weight factor as $\lambda=0.5$.
\subsubsection{Baselines as Alternate Fine-tuning approaches.}

\textbf{Token Weight Cross Entropy (TW-CE).}
The idea of TW-CE appeared original in dense object detection \cite{lin2018focallossdenseobject}. In this work, they provide a weight for each category to measure the cross-entropy function. The intuition of this weight assignment is that some categories have higher chance to be selected compared to other category in object detection. We borrow this idea to implement the baseline with a token weight applied to the original cross-entropy. 

We implement the cross-entropy loss by reweighting individual target tokens according to their structural importance in the power outage report. The first step of TW-CE created a fixed vocabulary-level weight vector by decoding each token and assigning higher weights to tokens indicative of XML structure (e.g., tag markers such as \texttt{<} and \texttt{>}), while leaving ordinary content tokens at uniform weight. During training, it computes the match between tags with higher score than the match between normal tokens. This preserves the efficiency and stability of CE while prioritizing learning on XML-critical tokens.
% , following the same general principle as classic cost-sensitive weighting and modern reweighting variants of cross-entropy used to emphasize difficult or important training signals~\cite{lin2017focal,zadrozny2003cost,cui2019classbalanced}.

\textbf{Tag Masking Cross Entropy (TM-CE).}
The idea of Tag Masking for cross-entropy in fine-tuning models has come from the BERT-based model \cite{devlin2019bertpretrainingdeepbidirectional,joshi2020spanbertimprovingpretrainingrepresenting}. In the BERT model, given input as a sequence of tokens, a set of tokens was replaced by a masking token. During the training process, the masking tokens will be predicted, thus improving the quality of models for text generation. SpanBERT \cite{joshi2020spanbertimprovingpretrainingrepresenting} also relied on masking technique, but this work attempted to prioritize masking spans of text instead of masking discontinuous tokens. Usually, the masked tokens/ spans were considered as important content inside the input for training models.

For power outage report generation, we implement a module that injects XML-structure awareness into standard causal language modeling by requiring the model to identify which response tokens belong to XML tags (i.e., substrings matching \texttt{<\ldots>} within the response region) and upweighting their contribution to the cross-entropy in fine-tuning. Our implementation of tag-masking CE concatenates the prompt and response into a single sequence, locates the response span using a delimiter, detects all tag character spans via a regular expression, and uses a fast tokenizer’s offset mapping to project these character spans onto token indices to construct a binary \texttt{tag\_mask}. The training labels follow the usual instruction-tuning convention where prompt tokens are ignored and only response tokens contribute to the loss. During optimization, the trainer computes per-token cross-entropy under the standard causal shift and applies a per-token weight of $w_{\text{tag}}$ to tokens with \texttt{tag\_mask}=1 and $w_{\text{text}}$ otherwise; an optional normalization rescales the weights so that the overall loss magnitude remains comparable to vanilla CE, avoiding unintended learning-rate effects. This is closely related in spirit to objectives that concentrate supervision on selected tokens or spans while keeping a CE-style optimization pipeline \cite{devlin2019bertpretrainingdeepbidirectional,joshi2020spanbertimprovingpretrainingrepresenting}.

\textbf{Example Weight Cross Entropy with PO Custom Loss (EW-CE-PO).}
This pipeline was implemented similarly to the Token Weight Cross Entropy, but the weight was calculated per sample instead of per token for cross-entropy calculation.This implementation was inspired from the success of reweighting examples for deep learning applications \cite{ren2019learningreweightexamplesrobust,shu2019metaweightnetlearningexplicitmapping}. We assign the weight as POTracker combined metric between labels and predicted results for reweighting example for cross entropy calculation.

\textbf{Direct Preference Optimization (DPO).}
Preference optimization \cite{furnkranz2010preference} is a technique aimed at aligning language models with human preferences by training them on datasets where outputs are ranked based on desirability. 
% Traditional methods, such as Reinforcement Learning from Human Feedback (RLHF), involve training a reward model to predict preferences as standard power outage reports that humans accepted and then fine-tuning the language model to maximize these predicted rewards. However, RLHF can be complex and computationally intensive due to the need for explicit reward modeling and reinforcement learning algorithms. To address these challenges, Direct Preference Optimization (DPO) \cite{rafailov2023dpo} has been proposed as an alternative approach. DPO simplifies the alignment process by directly optimizing the language model using preference data, eliminating the need for a separate reward model or reinforcement learning loop. 
% It achieves this by formulating a contrastive objective that increases the likelihood of preferred outputs over less preferred ones, effectively treating the alignment task as a supervised learning problem over preference pairs. This method has been shown to be stable, performant, and computationally efficient, matching or exceeding the performance of traditional RLHF approaches in aligning language models with human preferences.

In our implementation, there is one challenge that is different from other fine-tuning approaches: Direct Preference Optimization requires the training and validation set to have examples of bad responses besides good responses. From power outage reports in nonstandard and standard formats, we construct responses as follows. Given each gold XML response, the code generates a negative example (\texttt{bad\_response}) by attempting to parse the XML and applying one random corruption strategy: swap text between two leaf nodes; replace a leaf text with an implausible value; remove a random element, or (iv) perturb numeric leaf values. If parsing fails or the corrupted output is too similar/too short,a fallback routine truncates the XML and injects broken/incorrect tags. 

% Separation
% \textbf{Minimum Risk Training (MRT).}
% While other approaches of fine-tuning have more applications in text generation, Minimum Risk Training (MRT) is a traditional fine-tuning approach for sequence-to-sequence machine translation between natural languages. Since our approach can also be considered as the translation between domain-specific languages, we implement a Minimum Risk Training pipeline for power outage report generation following the idea provided by the work of Shen et al \cite{shen-etal-2016-minimum}. The advantage of MRT is provided by the ability to replace token-level maximum likelihood with minimizing the loss of full paragraph prediction under its list of n-best candidates for each time generation. We keep the idea of using BLEU score \cite{papineni-etal-2002-bleu} for loss calculation for our pipeline using MRT for power report generation.

\subsubsection{Baseslines as Rule-based Approaches.}
Besides LLM-based approaches, an ad-hoc but simpler approach for XML generation can be using a rule-based mechanism for power outage report generation. The main idea of this approach is provided as follows. First, we collect a set of rules for converting non-standard user input report tags to standard output tags. Next, we implement a program for traversing the input code, and heuristically convert each input tag to a standard output tag. The program can leave as it visits a tag that hasn't appeared in the vocabulary of rules we defined. In fact, this rule-based approach is inspired from research works on XML conversion in other languages or domains \cite{10.1007/978-3-540-39403-7_26,10.1007/978-3-540-69611-7_19}. 

One challenge of conducting rules for power outage tracker generation is that they need to be collected from experts. To achieve that set of rules, we hire two data scientists who have worked with energy sector datasets for more than 2 years. They manually look at 100 samples of non-standard and standard outage reports and write rules. Next, they discussed, among other about, the validity of each rule. The rules agreed upon by both were included in the implementation. The subset of rules 
% as frequent rules appeared from nonstandard to standard XML conversion 
is shown in the Appendix.

% In the era of LLMs, one challenge is that the proposed LLM has to be adapted to new knowledge. To address this, multiple pipelines for fine-tuning the LLMs to do a specific task have been proposed. The common idea of fine-tuning pipelines for LLMs is that they both rely on training and validation datasets with the following information: sample prompt and expected responses provided for the form. After the fine-tuning process, a set of weighted adapters was generated to direct original LLMs answering domain-specific questions. In our work, beside our proposed pipelines, we also perform two other branches of baselines. First, we implemented advantages of other fine-tuning approaches and used the generated adapters to evaluate the quality of power report generation. Second, we implemented a simpler rule-based conversion approach for power report generation for comparison with our work.

%% file: algms/alg_potracker_customloss.tex
\begin{algorithm}[t]
\caption{POTracker Custom Loss Function}
\label{alg:potracker-customloss}
\begin{algorithmic}[1]
\REQUIRE logits $Z$, labels $Y$ (prompt masked with $-100$), similarity metric $\mathrm{Sim}(\cdot,\cdot)$, weight $\lambda$
\ENSURE $\mathcal{L}_{\text{total}}$
\STATE Compute token-level cross-entropy on response tokens: $\mathcal{L}_{\text{CE}} \leftarrow \mathrm{CE}(Z, Y)$
\STATE Greedily decode a predicted response $\hat{y}$ from $Z$ and decode reference response $y$ from $Y$
\STATE Compute sequence-level quality score $s \leftarrow \mathrm{Sim}(y,\hat{y})$ \hfill (text + XML/tag structure)
\STATE Convert to penalty $p \leftarrow 1 - s$
\STATE Return total loss: $\mathcal{L}_{\text{total}} \leftarrow \mathcal{L}_{\text{CE}} + \lambda\, p$
\end{algorithmic}
\end{algorithm}

%% file: Experiment.tex
\section{Experiment}
In our experiments, we evaluate power outage report generation across multiple fine-tuning strategies and a rule-based conversion baseline. Next, we describe the evaluation setup.
% % Research Questions (RQ)
% \begin{enumerate}
%     \item RQ1. What is the accuracy of POTracker in Power Outage Report Generation?
%     \item RQ2. How number of parameters affect the accuracy of Power Outage Report Generation?
%     \item RQ3. How do textual similarity and tag similarity as factors affect the quality of POTracker?
%     \item RQ4. Can rule-based techniques replace LLMs in Power Outage Report Generation?
%     \item RQ5. Alternate strategies for implementing the POTracker custom loss function for the fine-tuning process. 
% \end{enumerate}

% In both RQs, we use our proposed $POTracker\_Loss$ for evaluating the quality of generated XMMs as PO Reports.
\subsection{Datasets Preparation}
% \subsubsection{Datasets for PO Reports Standardization}
We collect the power outage reports from Janurary, 2024 to March, 2025. We select 1000 data points for standard report generation using GPT-5.2. In this set, we randomly select 100 data points for evaluation. We manually verified this evaluation set to see if it has meaningful report information. For the remaining 900 data points we use them as the training data for the fine-tuning algorithms. In total, we use 900 data points for LLM Optimization techniques and use 100 data points for evaluation. We create a compact prompt with the integration of an example of standard power outage reports as context to help GPT-5.2 generate results with high quality. We perform multiple queries using the OpenAI key and the OpenRouter interface to achieve this dataset.

\subsection{Settings of Open LLM}

Qwen-2.5-7B-Instruct provides a strong accuracy--efficiency trade-off for domain assistants that must follow instructions and emit schema-constrained outputs. Despite its compact size, the model is explicitly optimized for instruction following and structured generation---capabilities that transfer well to XML-style report synthesis~\cite{qwen25modelcard,yang2025qwen25}. It also supports long-context inference (up to 128K tokens) and long-form generation (up to 8K tokens), which is beneficial when prompts include extensive context, guidelines, or multi-record outage narratives~\cite{qwen25modelcard}. In addition, Qwen2.5 models (including the 7B variant) are released as open weights under a permissive license (Apache 2.0 for all but specific excluded sizes), enabling reproducible research and practical deployment in real applications~\cite{qwen25blog}.

Given Qwen-2.5-7B-Instruct as our model for implementing strategies of fine-tuning approaches, we use popular hyperparameter settings for this model. It uses 28 Transformer layers with grouped-query attention (28 query heads and 4 key/value heads), Rotary Positional Embedding (RoPE) positional encoding, and Root Mean Square Layer Normalization (RMSNorm), and is released with bfloat16 weights. For text generation, we use the released default sampling setup enables stochastic decoding (\texttt{do\_sample}=true) with moderate temperature and nucleus/top-$k$ filtering, alongside a mild repetition penalty. Important setting of Qwen-2.5-7B-Instruct can be shown in the Appendix. 

\textbf{Prompt for answer generation.}
For all configurations except label generation, we do not include in-context examples of standard power outage reports. This design isolates the effect of fine-tuning by encouraging the model to learn the report format and domain conventions from the training and validation data, rather than relying on few-shot prompting or contextual imitation during inference.

\textbf{Original Baseline}. In the experiment, we add an extra configuration that generates a power outage report with the original selected LLM. For prompt preparation with this setting, we provide a hint for the report generation process that the output must follow the CIM-IEC-61968-3 format.

\subsection{Hardware Setting}
We ran all training and evaluation on a dedicated single-node server equipped with \texttt{1}$\times$\texttt{NVIDIA A100} GPU (\texttt{40},GB VRAM). The host system uses an \texttt{Intel Core i9} CPU with \texttt{16} physical cores and \texttt{100},GB RAM, backed by a \texttt{1},TB high-throughput NVMe SSD. Experiments were executed on \texttt{Ubuntu 24.04} with CUDA \texttt{12.0}.
\subsection{Evaluation Metrics}
We evaluate the quality of power report generation by the combined accuracy, text accuracy, and tag accuracy measured by $POTracker_{Metric}$. Noted, the text accuracy score was calculated when the weight of $POTracker_{Metric}$ was 1, and the tag accuracy score was calculated when the weight as 0 (see Formula \ref{eq:overall_similarity}).

\subsection{Results}
\input{tables/tbl_rq1}

\paragraph{Original LLM limitations.}
Table~\ref{tab:config_results} indicates that the baseline \textit{Qwen2.5-7B-Instruct} performs poorly for outage-report XML generation (Overall $16.20\%$, Tag Accuracy $3.56\%$). This result is consistent with the intuition that the target outputs are highly \emph{domain- and schema-specific}: generating standard-compliant XML requires strict adherence to hierarchical structure, tag ordering, and attribute constraints. While the baseline model can often produce plausible natural language, it struggles to consistently emit valid, standardized XML without additional task-specific supervision.

\paragraph{Cross-entropy variants do not transfer cleanly.}
The three cross-entropy fine-tuning variants (Token Weight, Tag Mask, and Example Weight) yield no meaningful gains over the baseline (Overall $\approx 15.87\%$--$16.13\%$, Tag Accuracy $\approx 3.48\%$--$3.50\%$). This suggests that standard NLP fine-tuning pipelines require adaptation when applied to power outage tracker report generation, where success depends primarily on \emph{structured correctness} rather than surface-level fluency. Reweighting tokens, masking tags, or upweighting examples may be insufficient when local token likelihood is weakly correlated with structural validity and schema compliance.

\paragraph{Preference optimization depends on negative quality.}
Direct Preference Optimization (DPO) remains close to the cross-entropy baselines (Overall $16.16\%$, Tag Accuracy $3.43\%$), implying that the preference signal may not yet be informative enough. A likely explanation is that negative example generation needs improvement: if negatives are too trivial, too noisy, or not targeted toward \emph{near-miss} structural failures, the model receives limited guidance about which corrections matter. Constructing harder negatives that preserve semantic plausibility while violating schema constraints (e.g., missing wrappers, invalid attributes, swapped tag order) may yield a stronger ranking signal and better alignment.

% \paragraph{MRT is effective but more expensive.}
% Minimum Risk Training (MRT) achieves strong performance (Overall $68.95\%$, Tag Accuracy $87.17\%$), comparable to the proposed POTracker Loss (Overall $69.14\%$, Tag Accuracy $86.47\%$). However, MRT requires substantially more compute in fine-tuning, taking $1437$ seconds compared to $772$ seconds for POTracker Loss (approximately $1.86\times$ slower). This highlights a practical trade-off: MRT can provide excellent task-level optimization, but POTracker Loss offers a more efficient accuracy--runtime balance for iterative development and scaling.

\paragraph{Rules remain a strong structural prior.}
Rule-based conversion (Overall $61.46\%$, Tag Accuracy $79.00\%$) still outperforms the LLM-only cross-entropy and DPO baselines by a wide margin. This reinforces that deterministic rules excel at enforcing structural constraints even when they lack semantic flexibility. 
% In future work---especially when extending to other domain datasets---a promising direction is to integrate rules as scaffolding for LLMs (e.g., rule-generated XML skeletons, schema validators, and post-hoc repair), enabling the model to focus on content completion while the system preserves schema compliance and interoperability.
\subsection{Analysis on quality of Ground-Truth Labels}
To assess the quality of the generated labels, we recruited two domain experts to manually evaluate 50 samples randomly selected from the POTracker dataset. Each expert was asked to assign a score from 1 to 5, where 1 indicates the lowest quality and 5 indicates the highest quality. The evaluation rubric was designed to measure the meaningfulness and validity of the generated data points. Based on this analysis, the generated labels achieved an average expert rating of \textbf{4.03} across the 100 randomly selected data points, with a median score of \textbf{5}. We provide the detailed analysis results in the Appendix and supplemental materials.

%% file: tables/tbl_rq1.tex
\begin{table}[t]
\vspace{-0.05in}
\caption{Performance comparison across training objectives and baselines.}
\label{tab:config_results}
\centering
\small
\begin{tabular}{lccc}
\toprule
\textbf{Configuration} & \textbf{Overall} & \textbf{Text Acc/} & \textbf{Tag Acc.} \\
\midrule
Qwen2.5-7B-Instruct      & 16.20\% & 28.85\% & 3.56\% \\
TW-CE         & 15.93\% & 28.39\% & 3.48\% \\
TM-CE              & 16.13\% & 28.76\% & 3.50\% \\
EW-CE-PO        & 15.87\% & 28.25\% & 3.49\% \\
DPO     & 16.16\% & 28.89\% & 3.43\% \\
Rule-based Conversion               & 61.46\% & 43.91\% & 79.00\% \\
% MRT         & 68.95\% & 50.73\% & 87.17\% \\
POTracker Loss                      & 69.14\% & 51.81\% & 86.47\% \\
\bottomrule
\end{tabular}
\vspace{-0.10in}
\end{table}

%% file: RelatedWork.tex
\section{Related Work} 
% Recent advances in large language models (LLMs) have driven substantial progress in both structured data understanding and structured output generation. 
Pretrained architectures such as BERT~\cite{devlin2019bert}, CodeT5~\cite{wang2021codet5}, and LLaMA~\cite{touvron2023llama} are now widely used for tasks requiring rich contextual representations and fine-grained prediction. Supervised fine-tuning (SFT) is a strong and practical approach for adapting these models to domain-specific settings~\cite{parthasarathy2024ultimateguidefinetuningllms}, including low-resource scenarios and cases where training labels are synthetically constructed~\cite{mecklenburg2024injectingnewknowledgelarge}. Beyond SFT, preference optimization techniques—such as Reinforcement Learning from Human Feedback (RLHF)~\cite{ouyang2022training} and Direct Preference Optimization (DPO)~\cite{rafailov2023dpo}—have been shown to improve alignment with human preferences in text generation. 

%% file: Discussion.tex
\section{Limitations}
While the POTracker framework demonstrates promising results in structured report generation, several limitations should be acknowledged. First, our study focuses exclusively on \textit{domain-specific problems} related to power outage tracking, and does not generalize to related domains such as weather forecasts. Second, the fine-tuning process relies heavily on \textit{synthetic datasets} generated via prompting and weak supervision, which may introduce alignment artifacts or limit diversity compared to real-world annotated data. 
% Third, our evaluation was conducted using a fixed model size LLM as Qwen-2.5-7B-Instruct; we \textit{did not explore multiple parameter scales} (e.g., 7B vs. 13B vs. 70B), leaving open questions about how model capacity influences performance. Finally, the current system is \textit{designed and evaluated solely in English}, and does not yet support multilingual report generation, which is critical for global deployment. 
% These constraints offer avenues for future research on domain transferability, real-world data integration, model scaling, and cross-lingual capabilities.

%% file: Conclusion.tex
\section{Conclusion}
We introduce \textit{POTracker}, a domain-specific AI assistant that supports power outage tracking via LLM-based reasoning. POTracker standardizes heterogeneous, free-form outage reports into ODIN-compliant XML (CIM-IEC-61968-3), with a focus on improving the reliability of compact open LLMs for high-precision structured generation. We incorporate multiple adaptation strategies, including supervised fine-tuning and preference-based optimization via Direct Preference Optimization (DPO), and compare against a rule-based conversion baseline. Across our experiments, POTracker improves over the base LLM by up to 53\% of overall accuracy. 
% Our best configuration achieves 51.81\% textual accuracy and 87.17\% tag accuracy, yielding a combined score of 69.14\% under $POTracker_{Metric}$. 
% Overall, the results show that injecting domain knowledge, leveraging curated context, and applying preference-based alignment substantially enhances LLM performance for real-world, schema-constrained applications in energy infrastructure monitoring.

% \section*{Acknowledgments}
% We would like to thank the scientists and domain experts at Oak Ridge National Laboratory (ORNL) for their invaluable support in curating and validating the power outage datasets used in this study. Their expertise was essential in generating high-quality standard reports and providing domain-specific insights that guided the design and evaluation of the POTracker system.

%% file: Acknowledgement.tex
\section*{Acknowledgments}
We acknowledge that this manuscript has been authored by UT-Battelle, LLC under Contract No. DE-AC05-00OR22725 with the U.S. Department of Energy. The United States Government retains and the publisher, by accepting the article for publication, acknowledges that the United States Government retains a non-exclusive, paid-up, irrevocable, world-wide license to publish or reproduce the published form of this manuscript, or allow others to do so, for United States Government purposes. DOE will provide public access to these results of federally sponsored research in accordance with the DOE Public Access Plan (http://energy.gov/downloads/doe-public-access-plan). Research sponsored by the Laboratory Directed Research and Development Program of Oak Ridge National Laboratory, managed by UT-Battelle, LLC, for the U. S. Department of Energy.

%% file: Appendix.tex
\section{Appendix}
\subsection{Algorithm of Token Weight Cross Entropy}

% Preamble (ICML typically supports these):
% \usepackage{algorithm}
% \usepackage{algpseudocode}

\begin{algorithm}
\caption{Token-Weighted Cross Entropy Loss (Causal LM)}
\label{alg:token_weight_ce_loss}
\begin{algorithmic}[1]
\REQUIRE Logits $\mathbf{Z}\in\mathbb{R}^{B\times T\times V}$, labels $\mathbf{Y}\in\{-100,0,\dots,V-1\}^{B\times T}$
\REQUIRE Vocabulary weights $\mathbf{w}^{(v)}\in\mathbb{R}^{V}$, ignore index $-100$, normalize flag $norm$
\ENSURE Scalar loss $\mathcal{L}$
% \FUNCTION{TokenWeightedCE}{$\mathbf{Z}, \mathbf{Y}, \mathbf{w}^{(v)}, norm$}
    \STATE $\mathbf{Z}' \gets \mathbf{Z}_{:,\;1:T-1,\;:}$ \COMMENT{SHIFT LOGITS LEFT}
    \STATE $\mathbf{Y}' \gets \mathbf{Y}_{:,\;2:T}$ \COMMENT{SHIFT LABELS RIGHT}
    \STATE $\mathbf{M} \gets \mathbb{I}[\mathbf{Y}' \neq -100]$ \COMMENT{VALID-TOKEN MASK, SHAPE $(B,T-1)$}
    \STATE $\boldsymbol{\ell} \gets \textsc{CrossEntropyNoReduce}(\mathbf{Z}', \mathbf{Y}', \texttt{ignore}=-100)$
    \COMMENT{$\boldsymbol{\ell}\in\mathbb{R}^{B\times (T-1)}$}
    \STATE $\mathbf{S} \gets \mathbf{Y}'$;\;\; $\mathbf{S}[\mathbf{S}=-100]\gets 0$ \COMMENT{SAFE GATHER INDICES}
    \STATE $\mathbf{W} \gets \mathbf{w}^{(v)}[\mathbf{S}] \odot \mathbf{M}$ \COMMENT{TOKEN WEIGHTS PER POSITION}
    \IF{$norm$}
        \STATE $d \gets \max\left(\sum \mathbf{W},\;1\right)$
        \STATE $c \gets \max\left(\sum \mathbf{M},\;1\right)$
        \STATE $\mathbf{W} \gets \mathbf{W}\cdot \frac{c}{d}$ \COMMENT{KEEP LOSS SCALE CE-LIKE}
    \ENDIF
    \STATE $c \gets \max\left(\sum \mathbf{M},\;1\right)$
    \STATE $\mathcal{L} \gets \frac{\sum (\boldsymbol{\ell}\odot \mathbf{W})}{c}$
    \STATE $\mathcal{L}$
% \ENDFUNCTION
\end{algorithmic}
\end{algorithm}

\subsection{Algorithm of Tag Mark Cross Entropy}

% ICML-ready pseudocode (single-column friendly)
% Preamble:
% \usepackage{algorithm}
% \usepackage{algpseudocode}

\begin{algorithm}
\caption{Tag-Mask Weighted Cross-Entropy (Tag Mask CE)}
\label{alg:tagmask-ce}
\begin{algorithmic}[1]
\REQUIRE Model logits $Z \in \mathbb{R}^{B \times T \times V}$, labels $Y \in \mathbb{Z}^{B \times T}$ (with $-100$ as ignore),
\REQUIRE tag mask $M \in \{0,1\}^{B \times T}$, weights $w_{\text{tag}}$, $w_{\text{text}}$, normalize flag $\textsc{Normalize}$
\ENSURE Scalar loss $L$

% \FUNCTION{ComputeTagMaskCELoss}{$Z, Y, M, w_{\text{tag}}, w_{\text{text}}, \TEXTSC{Normalize}$}
  \STATE $Z' \gets Z[:, 0{:}T{-}1, :]$ \COMMENT{$B \times (T{-}1) \times V$}
  \STATE $Y' \gets Y[:, 1{:}T]$         \COMMENT{$B \times (T{-}1)$}
  \STATE $M' \gets M[:, 1{:}T]$         \COMMENT{$B \times (T{-}1)$}

  \STATE $CE \gets \textsc{CrossEntropyNoReduce}(Z', Y', \textsc{ignore}=-100)$ \COMMENT{$B \times (T{-}1)$}

  \STATE $Vmask \gets \mathbb{I}[Y' \neq -100]$ \COMMENT{$B \times (T{-}1)$}

  \STATE $W \gets \big(M' \cdot (w_{\text{tag}} - w_{\text{text}}) + w_{\text{text}}\big) \odot Vmask$

  \IF{$\textsc{Normalize}$}
    \STATE $\textit{denom} \gets \max\big(\sum W,\ 1\big)$
    \STATE $\textit{target} \gets \max\big(\sum Vmask,\ 1\big)$
    \STATE $W \gets W \cdot (\textit{target} / \textit{denom})$
  \ENDIF

  \STATE $L \gets \dfrac{\sum (CE \odot W)}{\max(\sum Vmask,\ 1)}$
  \STATE $L$
% \ENDFUNCTION

\end{algorithmic}
\end{algorithm}

\subsection{Algorithm of Example Weight Cross Entropy}
% Example-weighted CE loss (as in fine-tuned_potracker_exampleweightce.py)
% Requires: \usepackage{algorithm} \usepackage{algpseudocode}

\begin{algorithm}
\caption{Example-Weighted Cross-Entropy Loss (POTracker-style)}
\label{alg:exampleweight-ce}
\begin{algorithmic}[1]
\REQUIRE logits $L \in \mathbb{R}^{B \times T \times V}$, labels $Y \in \mathbb{Z}^{B \times T}$ (ignore index $=-100$)
\REQUIRE tokenizer $\mathcal{D}$, \textsc{CombinedSimilarity}$(\cdot,\cdot,\alpha)$
\REQUIRE hyperparams: $\lambda$ (potracker\_weight), $N$ (potracker\_every), mode $\in\{\texttt{1-minus},\texttt{inv}\}$, $\epsilon$, clip range $[w_{\min}, w_{\max}]$, EMA factor $\beta$
\REQUIRE state: call\_count, EMA baseline $\mu$ (initialized as \texttt{None})
\ENSURE scalar loss

% \Function{ComputeLoss}{$L, Y$}
  \STATE $L' \gets L[:, 1{:}T, :]$;\ \ $Y' \gets Y[:, 2{:}T]$ \COMMENT{causal shift (teacher forcing)}
  \STATE $ce_{\text{tok}} \gets \textsc{CrossEntropy}(L', Y', \texttt{reduction=none}, \texttt{ignore}=-100)$ \COMMENT{$B\times(T{-}1)$}
  \STATE $m \gets \mathbb{I}[Y' \neq -100]$;\ \ $c \gets \max(1, \sum_{t} m)$ \COMMENT{valid-token mask/count per example}
  \STATE $ce_{\text{ex}} \gets \frac{\sum_{t} (ce_{\text{tok}} \odot m)}{c}$ \COMMENT{per-example CE, shape $(B,)$}

  \STATE call\_count $\gets$ call\_count $+ 1$
  \STATE do\_score $\gets$ (\texttt{call\_count mod} $N = 0$) $\wedge$ ($\lambda>0$)

  \IF{do\_score}
    \STATE $\hat{Y} \gets \arg\max_{v} L$ \COMMENT{token-level argmax predictions, shape $(B,T)$}
    \FOR{$i=1$ to $B$}
      \STATE $S_i \gets \{t \mid Y_{i,t} \neq -100\}$ \COMMENT{positions used for scoring}
      \STATE $y_i \gets \mathcal{D}(\,Y_{i,S_i}\,)$;\ \ $\hat{y}_i \gets \mathcal{D}(\,\hat{Y}_{i,S_i}\,)$ \COMMENT{decode to text/XML}
      \STATE $s_i \gets \textsc{CombinedSimilarity}(y_i,\hat{y}_i,\alpha)[\texttt{combined\_similarity}]$ \COMMENT{typically in $[0,1]$}
      \IF{mode == \texttt{inv}}
        \STATE $p_i \gets \frac{1}{s_i + \epsilon}$
      \ELSE
        \STATE $p_i \gets 1 - s_i$
      \ENDIF
    \ENDFOR

    \STATE $\bar{p} \gets \frac{1}{B}\sum_i p_i$
    \IF{$\mu$ is \texttt{None}} \STATE $\mu \gets \bar{p}$
    \ELSE \STATE $\mu \gets \beta\mu + (1-\beta)\bar{p}$ \ENDIF

    \STATE $w_i \gets \textsc{Clip}\big(1 + \lambda(p_i - \mu),\, w_{\min},\, w_{\max}\big)$
    \STATE $w \gets \textsc{StopGrad}(w)$ \COMMENT{detach weights from backprop}
  \ELSE
    \STATE $w \gets \mathbf{1}$ \COMMENT{no reweighting this batch}
  \ENDIF

  \STATE $\frac{1}{B}\sum_{i=1}^{B} \big(w_i \cdot ce_{\text{ex},i}\big)$
  \COMMENT{important: mean($w\cdot ce$), not normalized by $\sum w$}
% \EndFunction
\end{algorithmic}
\end{algorithm}

\subsection{Algorithm for bad response generation for DPO training.}
% Preamble (ICML-friendly):
% \usepackage{algorithm}
% \usepackage{algpseudocode}

\begin{algorithm}
\caption{High-Level Bad-Response Generation for DPO (XML Corruption)}
\label{alg:dpo_bad_response}
\begin{algorithmic}[1]
\REQUIRE Chosen XML string $x^{+}$, RNG $\mathcal{R}$, optional other chosen XML $x^{\text{other}}$
\ENSURE Rejected XML string $x^{-}$
% \FUNCTION{MAKEBADRESPONSE}{$x^{+}, \mathcal{R}, x^{\text{other}}$}
    \STATE $x^{+} \leftarrow \textsc{Trim}(x^{+})$
    \STATE $(ok, T) \leftarrow \textsc{TryParseXML}(x^{+})$
    \IF{$\neg ok$}
        \STATE \textsc{FallbackCorrupt}$(x^{+}, \mathcal{R})$
    \ENDIF

    \STATE $E \leftarrow \textsc{AllElements}(T)$
    \STATE $P \leftarrow \textsc{BuildParentMap}(T)$
    \STATE $L \leftarrow \textsc{LeafTextNodes}(T)$ \COMMENT{LEAVES WITH NONEMPTY TEXT}

    \STATE $O \leftarrow \emptyset$
    \IF{$x^{\text{other}} \neq \emptyset$}
        \STATE $(ok_2, T_2) \leftarrow \textsc{TryParseXML}(\textsc{Trim}(x^{\text{other}}))$
        \IF{$ok_2$}
            \STATE $O \leftarrow \textsc{LeafTexts}(T_2)$ \COMMENT{POOL OF PLAUSIBLE TEXT VALUES}
        \ENDIF
    \ENDIF

    \STATE $\mathcal{S} \leftarrow [\textsc{SwapLeafText}, \textsc{ReplaceLeafText}, \textsc{RemoveElement}, \textsc{NumericPerturb}]$
    \STATE $s \leftarrow \textsc{RandomChoice}(\mathcal{R}, \mathcal{S})$

    \IF{$s=\textsc{SwapLeafText}$ \AND $|L|\ge 2$}
        \STATE $(a,b) \leftarrow \textsc{RandomSample2}(\mathcal{R}, L)$
        \STATE \textsc{SwapText}$(a,b)$
    \ELSIF{$s=\textsc{ReplaceLeafText}$ \AND $|L|\ge 1$}
        \STATE $t \leftarrow \textsc{RandomChoice}(\mathcal{R}, L)$
        \IF{$|O|\ge 1$}
            \STATE $\textsc{SetText}(t,\ \textsc{RandomChoice}(\mathcal{R}, O))$
        \ELSE
            \STATE $\textsc{SetText}(t,\ \textsc{RandomChoice}(\mathcal{R}, \{\texttt{UNKNOWN},\texttt{N/A},\texttt{0},\texttt{999999},\texttt{Pending Investigation}\}))$
        \ENDIF
    \ELSIF{$s=\textsc{RemoveElement}$}
        \STATE $C \leftarrow \textsc{NonRootRemovable}(E, P)$
        \IF{$|C|\ge 1$}
            \STATE $v \leftarrow \textsc{RandomChoice}(\mathcal{R}, C)$
            \STATE \textsc{RemoveChild}$(P[v], v)$
        \ELSIF{$|L|\ge 1$}
            \STATE $\textsc{SetText}(\textsc{RandomChoice}(\mathcal{R}, L),\ \texttt{UNKNOWN})$
        \ENDIF
    \ELSIF{$s=\textsc{NumericPerturb}$ \AND $|L|\ge 1$}
        \STATE $t \leftarrow \textsc{RandomChoice}(\mathcal{R}, L)$
        \IF{\textsc{IsNumeric}(\textsc{GetText}(t))}
            \STATE $v \leftarrow \textsc{ToFloat}(\textsc{GetText}(t))$
            \STATE $v \leftarrow v + \textsc{RandomSign}(\mathcal{R}) \cdot \textsc{Uniform}(\mathcal{R}, 1, 1000)$
            \STATE $\textsc{SetText}(t,\ \textsc{FormatNumber}(v))$
        \ELSE
            \STATE $\textsc{SetText}(t,\ \textsc{RandomChoice}(\mathcal{R}, \{\texttt{0},\texttt{UNKNOWN},\texttt{999999}\}))$
        \ENDIF
    \ENDIF

    \STATE $x^{-} \leftarrow \textsc{SerializeXML}(T)$
    \STATE $x^{-} \leftarrow \textsc{EnsureXMLDeclaration}(x^{-})$

    \STATE $x^{+}_{decl} \leftarrow \textsc{EnsureXMLDeclaration}(x^{+})$
    \IF{$\textsc{Trim}(x^{-})=\textsc{Trim}(x^{+}_{decl})$ \OR $\textsc{Length}(x^{-})<60$}
        \STATE $x^{-} \leftarrow \textsc{FallbackCorrupt}(x^{+}, \mathcal{R})$
    \ENDIF
    \STATE $x^{-}$
% \ENDFUNCTION
\end{algorithmic}
\end{algorithm}

\section{Rule-based approach}
Table \ref{tab:xml-mapping-pcols} shows the subset of rules we use for the rule-based approach for converting from non-standard to standard outage report.

\input{algms/alg_rulebased}

\section{Settings for Qwen-2.5-7B-Instruct in POTracker's experiment}
Table \ref{tab:qwen25_7b_instruct_gen_defaults} shows the parameters we set as default for POTracker's experiment.

\begin{table}[t]
\caption{Settings for Qwen-2.5-7B-Instruct in POTracker's experiment}

\centering
\small
\begin{tabular}{ll}
\toprule
\textbf{Generation Parameter} & \textbf{Default} \\
\midrule
\texttt{do\_sample} & \texttt{true} \\
\texttt{temperature} & 0.7 \\
\texttt{top\_p} & 0.8 \\
\texttt{top\_k} & 20 \\
\texttt{repetition\_penalty} & 1.05 \\
\bottomrule
\end{tabular}
\label{tab:qwen25_7b_instruct_gen_defaults}
\end{table}

\section{Human Study on Quality of Ground-truth Labels}
\begin{table}[t]

\centering
\small
\caption{Distribution of human rating scores for generated ground-truth standard reports. Domain experts rated each label on a 0--5 scale.}
\begin{tabular}{lc}
\toprule
\textbf{Human Rating Score} & \textbf{Number of Data Points} \\
\midrule
0 & 1 \\
1 & 0 \\
2 & 12 \\
3 & 20 \\
4 & 16 \\
5 & 51 \\
\midrule
Average & \textbf{4.03} \\
Median & \textbf{5} \\
\bottomrule
\end{tabular}

\label{tab:human_label_quality}
\end{table}

\section{Accuracy on varying weight for POTracker}
We adjust the weight of POTracker\_Loss as $\alpha$ from 0 to 1 and see the change on overall accuracy. We get the accuracy as of POTracker as shown in Table \ref{tab:weight_change_potracker}.

\begin{table}[t]
\centering
\small
\caption{Accuracy of POTracker under different values of the weighting factor $\alpha$.}
\label{tab:weight_change_potracker}

\begin{tabular}{lc}
\toprule
\textbf{$\alpha$} & \textbf{Accuracy} \\
\midrule
0.1 & 83.00\% \\
0.2 & 79.54\% \\
0.3 & 76.07\% \\
0.4 & 72.60\% \\
0.5 & 69.14\% \\
0.6 & 65.67\% \\
0.7 & 62.21\% \\
0.8 & 58.74\% \\
0.9 & 55.27\% \\
1.0 & 51.81\% \\
\bottomrule
\end{tabular}
\end{table}

%% file: algms/alg_rulebased.tex
\begin{table}[t]
\centering
\small
\caption{Set of rules for conversion from non-standard to standard outage report}
\label{tab:xml-mapping-pcols}
\setlength{\tabcolsep}{6pt}
\renewcommand{\arraystretch}{1.15}
\begin{tabular}{|p{0.4\linewidth}|p{0.5\linewidth}|}
\hline
\centering\textbf{Nonstandard outage report} & \centering\textbf{Standard outage report} \tabularnewline
\hline
Outage/mRID & \texttt{<ns0:mRID>OUTAGE- \{area\}-0001</ns0:mRID>} \tabularnewline \hline
Outage/communityDescriptor & \texttt{<ns0:communityDescriptor> \{area\}</ns0:communityDescriptor>} \tabularnewline \hline
Outage/metersAffected & \texttt{<ns0:metersAffected> \{metersAffected\} </ns0:metersAffected>} \tabularnewline \hline
Outage/EstimatedRestorationTime/ert & \texttt{<ns0:ert>\{etr\}</ns0:ert>} \tabularnewline \hline
Outage/OutageArea/metersServed & \texttt{<ns0:metersServed>\{metersServed\} </ns0:metersServed>} \tabularnewline \hline
Outage/OutageArea/outageAreaKind & \texttt{<ns0:outageAreaKind>zipcode </ns0:outageAreaKind>} \tabularnewline \hline
Outage/Incident/cause & \texttt{<ns0:cause>Pending Investigation</ns0:cause>} \tabularnewline \hline
Outage/Incident/Location/ geoInfoReference & \texttt{<ns0:geoInfoReference>\{area\} </ns0:geoInfoReference>} \tabularnewline \hline
Outage/Incident/Location/zoneKind & \texttt{<ns0:zoneKind>zipcode< /ns0:zoneKind>} \tabularnewline \hline
\end{tabular}

\end{table}